# Conflict Detection in IoT-based Smart Homes

Bing Huang
School of Computer Science
The University of Sydney
Sydney, Australia
bing.huang@sydney.edu.au

Hai Dong
School of Computer Science
RMIT University
Melbourne, Australia
hai.dong@rmit.edu.au

Athman Bouguettaya
School of Computer Science
The University of Sydney
Sydney, Australia
athman.bouguettaya@sydney.edu.au

*Abstract*—We propose a novel framework that detects conflicts in IoT-based smart homes. Conflicts may arise during interactions between the resident and IoT services in smart homes. We propose a generic knowledge graph to represent the relations between IoT services and environment entities. We also profile a generic knowledge graph to a specific smart home setting based on the context information. We propose a conflict taxonomy to capture different types of conflicts in a *single* resident smart home setting. A conflict detection algorithm is proposed to identify potential conflicts using the profiled knowledge graph. We conduct a set of experiments on real datasets and synthesized datasets to validate the effectiveness and efficiency of our proposed approach.

## I. INTRODUCTION

The Internet is evolving from connecting computers to connecting *things*. The rapid development of Internet of Things (IoT) technology empowers *things* with capabilities such as sensing, actuating, and communication [1]. The connected things have functional and non-functional features. The key challenge is that connected things are highly heterogeneous in terms of supporting infrastructure ranging from networking to programming abstraction [2]. Service-oriented Computing (SOC) is a promising solution for abstracting connected *things* on the Internet as services by hiding the complex and diverse supporting infrastructure [3]. Such abstracted *things* on the Internet are referred to as IoT services. An example of IoT services is a light service whose functionality includes turning on and turning off and non-functionality(i.e., QoS) may include price.

An important application domain for IoT is smart homes. A smart home is any regular home that has been augmented with a variety of IoT services [4]. The goal of smart homes is to improve resident life *convenience* via reducing residents' interactions with IoT services. A typical scenario of life *convenience* is that the blind is pulled down automatically when it is dazzling outside in summer. Invoking IoT services on behalf of residents automatically is a promising solution to make residents free from repetitive efforts of interacting with IoT services. There are different ways of automating IoT services to fulfil residents' *convenience* needs, i.e., recommending and automating IoT services that the resident wants to use or developing ECA rules (Event-Condition-Action), also known as trigger-action rules, to invoke IoT services automatically [5]. The Event-ConditionAction paradigm is a promising approach because of its compact and intuitive structure, which directly connects the dynamic events and/or conditions (i.e., referred to as triggers) with the expected reactions (i.e., actions) without requiring the use of complex programming structures [6]. An ECA rule could be defined by a set of IoT services chained by IF-THEN structure and logical operators (i.e., AND and OR). An example of such a rule is that " If I arrive at home then turns on the TV".

With an increasing number of ECA rules developed by the residents, running multiple rules simultaneously may cause conflicting situations. *Conflicts* refer to situations resulting from the interference between IoT services. Conflicts usually cause undesirable situations and should be detected and resolved. There are two main categories of conflicts: *direct conflicts* and *indirect conflicts*. A direct conflict occurs when different ECA rules compete to invoke the shared IoT services at the same time. An example of direct conflicts is that an ECA rule requests to turn on the light while another requires to turn off the same light at the same time. An indirect conflict describes the situation that an IoT service may interfere with another IoT service in a subtle and implicit way due to their impacts on a shared environment entity. A typical example of indirect conflicts is that the AC is cooling while the window is opened (i.e., An opening window would let hot air come into the room and decrease the cooling effect in hot summer).

There are a few studies on detecting conflicts [7] [8] [9][10] [11] [12]. In [7], a framework RemedIoT is proposed to detect conflict among IoT-based applications based on a set of defined policies. In [8], conflicts are detect based on an object-oriented approach. Appliances and environment entities are modeled as a set of methods and properties. A conflict management system is developed to identify and resolve conflicts at system design stage [9]. In [10], a formal rule model called UTEA is proposed to represent a rule constituting user, triggers, environment entities, and actuators. Various types of conflicts among rules are detected through associating the effects of actuators with physical environment entities. In [11], an ontology-based approach is proposed to detect different types of conflicts among IoT services in multiple-residents smart home settings. A fine grained approach is proposed in [12] to further improve the ontology-based approach.

These work highly rely on expert annotations because of the lack of knowledge regarding the IoT services' impact on the environment. In this regard, These approaches are time consuming and labor intensive. To the best of our knowledge, there is little research on detecting *indirect* conflicts hidden among IoT services because of the lack of IoT services' information about their impact on the environment. A typical example of *indirect* conflict is that the AC is cooling while the window is opened (i.e., opening window would let hot air come in to the room and decrease the cooling effect). In practice, an IoT service may interfere with another IoT service in a subtle and implicit way due to their impacts on a shared environment property (e.g., AC vs window). Therefore, this paper focuses on

detecting both *direct* and *indirect* conflicts without expert annotations and in-

- We validate the effectiveness and efficiency of our proposed approach on real datasets and synthesized datasets.

TABLE I
EXAMPLES OF ECA RULES

| ECA rules | Rule description |
| --- | --- |
| R1: temperature > 25C → close window | If temperature is more than 25C, close the window. |
| R2: CO2 > 0.5% → open window | If carbon dioxide is more than 0.5%, open the window. |
| R3: weather = sunny → pull up blind | If it is sunny, pull up the blind. |
| R4: turn on TV → brightness = 50 | If TV is turned on, set brightness to be 50. |
| R5: temperature < 23C → turn on heater AND temperature =26C | If temperature is less than 23C, turn on the heater and set temperature to be 26C. |
| R6: temperature > 25C → turn on AC AND temperature = 24C | If temperature is more than 25C, turn on the AC and set temperature to be 24C. |
| R7: humidity > 50% → open window | If humidity is more than 50%, open the window. |
| R8: turn on stove → turn on kitchen hood | If the stove is turned on, turn on the kitchen hood. |

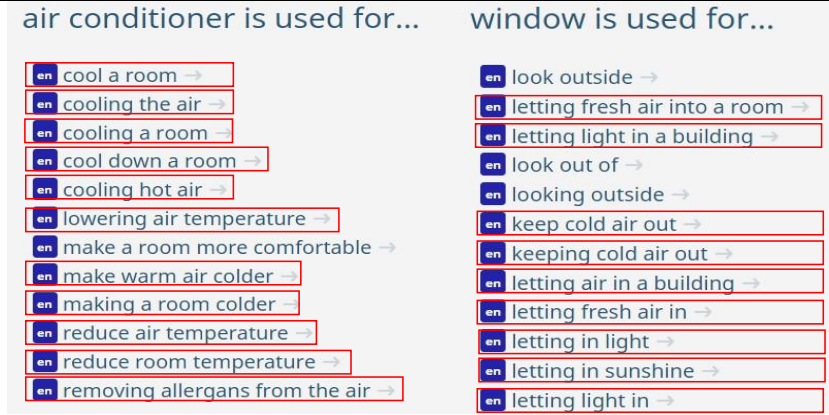

Fig. 1. Examples of IoT services on ConceptNet

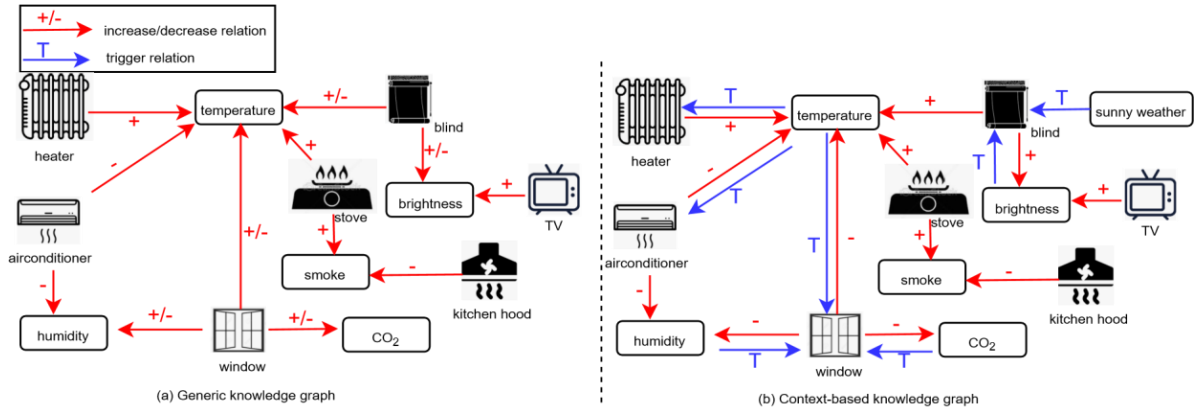

Fig. 2. (a). Generic knowledge graph and (b).Real-time context-based knowledge graph based on ECA rules and real-time environment variables

terventions. Specifically, we propose to employ NLP techniques to extract knowledge regarding IoT services from knowledge graphs and detect conflicts hidden among IoT services without experts' annotations. In a nutshell, the contributions of the paper are as follows.

- We propose a generic knowledge graph model to capture the relations between IoT services and environment properties. We also tailor the generic knowledge graph to a contextbased knowledge graph using a variety of contextual information.
- We propose a *conflict taxonomy* to define different types of conflicts.
- We propose a new conflict detection algorithm to detect *direct* and *indirect* conflicts.

## MOTIVATING SCENARIO

We discuss several motivating scenarios to illustrate different types of conflicts in a single resident smart home setting. Let us assume that a resident developed 8 ECA rules for his/her smart home using the IFTTT[1] programming tool as shown in TABLE.I. We identify eight common environment properties including *temperature, weather, brightness, humidity, smoke, sound, humidity, ventilation, and $CO_2$*. We also assume all the involved IoT services including a heater, an AC, a window, a stove, a blind, a kitchen hood, a TV, and a stove are deployed in the same room (i.e., a single studio). Generic knowledge graph construction and profiling and conflict modelling are necessary tasks before detecting conflicts. We briefly describe the three tasks as follows.

Understanding how IoT services interfere with each other plays a key role in detecting conflicts hidden among IoT services.

---
[1] https://ifttt.com/home

However, lacking knowledge regarding the IoT service is the main hindrance to conflict detection. We introduce the knowledge graph (i.e., ConceptNet [2]) as a main source of information for extracting IoT services' information. The knowledge graph provides a rich source of information about IoT services represented by natural languages. This information is organized and structured based on different types of relations. For example, Fig.1 shows the information about an AC and a window based on the *isUsedFor* property on ConceptNet. In this paper, we use the *isUsedFor* and the *isCapableOf* properties as our source of IoT service information because the two properties contain the majority of descriptions regarding IoT service functionalities. By using natural language processing (NLP) techniques and a lexical database (i.e., WordNet [3]), We can automatically extract the information about IoT services and their impacts on the eight common environment properties. As the example shown in Fig.2(a), we can extract the information that an AC can decrease temperature and a window can decrease/increase temperature. The symbols "+" and "−" represent *increase* and *decrease*, respectively.

The generic knowledge graph, also known as *seed knowledge graph*, could be applicable to any smart homes. However, it has an ambiguous relation between an IoT service and an environment (e.g., window and temperature). This is because the knowledge regarding IoT services is provided by different people. In this regard, the seed knowledge graph should be tailored based on the context of a specific smart home and real-time environment variable value. The ECA rules developed by residents provide a rich source of information regarding the smart home context. For example, if the real-time outdoor temperature is more than 25C and R1 is invoked, we can infer that the resident wants to close the window to prevent hot air. Therefore, we can decide the relationship between the window the temperature is "−" as shown in Fig.2. (b).

We can also extract *trigger* relations between environment properties and an IoT service from ECA rules. As the rule R1 described in TABLE I, the condition "temperature > 25C" may trigger a window to close. The ECA rules developed by residents provide a rich source of information to extract the relations between the environment properties and IoT services. We denote the trigger relations using the symbol $T$. As shown in Fig.2(b), *temperature* $-\rightarrow^T AC$ represents that the temperature change may invoke the AC to execute.

Once we tailor the seed knowledge graph by fixing *increase/decrease* relations and adding *trigger* relations based on the smart home's context, we obtain a profiled knowledge graph referred to as a real-time *context-based knowledge graph*. The context-based knowledge graph could serve as the knowledge base for detecting conflicts. The knowledge graph in Fig.2.(b) describes that how IoT services may interfere with each other via their impacts on the environment in a particular smart home based on the current environment variable values. In the rest of this paper, we define four types of conflicts based on the context-based knowledge graph.

*Function-function conflict*: This type of conflict describes the situation that two rules try to invoke a shared IoT service in an inverse manner. As illustrated in the rules R1 and R2, when their corresponding conditions are satisfied simultaneously (i.e., *temperature* > 25 and $CO_2$ > 0.5%), R1 requests to close the window while R2 requests to open the window. As a result, a *function-function conflict* occurs as the state of the window could not be opened and be closed at the same time.

*Cumulative-environment-impact conflict*: This type of conflict describes the situation that invoking two IoT service's functionality at the same time will impact the shared environment property, which is shown in the rules R3 and R4. If the rule R3 requests to pull up the blind while the resident is watching TV, the preferred brightness (i.e., 50) for watching TV may be impacted because pulling up the blind in sunny days would increase the indoor brightness. Consequently, the experience of watching TV becomes bad because of the dazzling sunlight. Another example of such conflict is shown between R5 and R8. It is possible that the temperature would be above the comfortable temperature 26C if the heater is heating and the stove is in use at the same time because using a stove would generate a lot of heat. As a result, the accumulative effect of simultaneously using heater and stove may increase the room temperature above a comfortable degree.

*Transitive-environment-impact-conflict*: This type of conflict describes the situation that the impact of an IoT service on the environment property would trigger the execution of another IoT service's functionality. As described in the rules R5 and R6, a *Transitive-environment-impact-conflict* may occur between the AC service and the heater service. If the temperature is 20C in cold winter, the heater service would start to heat the room temperature to 26C. As a result, the increasing temperature would trigger the AC service to cool the room to 24C.

*Opposite-environment-impact-conflict*: This type of conflict describes the situation that two IoT services change their shared environment property in an opposite way. An example of such conflict occurs between an AC service and a window service as shown in R6 and R7. When the AC is cooling the room with the window being opened, an *Opposite-environment-impact-conflict* arises because the opening window would increase the room temperature in hot summer while the AC tries to decrease the room temperature.

---

[2] https://conceptnet.io/
[3] https://wordnet.princeton.edu/

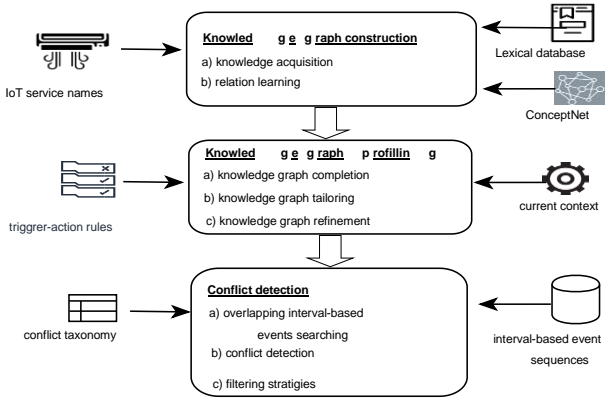

Fig. 3. Workflow of the proposed framework

## II. PRELIMINARY

We describe the notion of IoT service and ECA rules to model different types of conflicts. An IoT service is defined by functional and non-functional aspects. One or multiple IoT services are composed as a ECA rule to satisfy the residents' complex requirement. A conflict arises when multiple ECA rules are executed at the same time and place.

Definition 1: IoT service. An IoT service $S_i$ is represented by a tuple $S_i = <id, F_i, Q_i>$, where:

- $id$ is a unique IoT service identifier.
- $F_i = \{f_1, f_2...f_n\}$ is a set of functionalities offered by an IoT service.
- $Q_i = \{q_1, q_2...q_n\}$ is a set of IoT service qualities.

Definition 2: IoT service impact. An IoT service $S_i$ is considered to have an impact on an environment variable $env$ if invoking this IoT service can *increase/decrease* the value of the environment variable $env$. It is denoted as $imp = <S_i, eff, env>$ where $eff \in \{increase, decrease\}$. For example, the impact of the AC on temperature is denoted as "($AC, decrease, temperature$)" meaning that AC can decrease the temperature once it is invoked.

An IoT service may not satisfy the resident's complex requirement. Therefore, multiple IoT services can be composed to fulfil a resident' needs. These IoT services could be chained by simple language and logic operators. We refer such composed IoT services as a ECA rule. For example, IFTTT is a popular tool that provides residents with an intuitive programming environment to compose IoT services such as those examples shown in TABLE I. A ECA rule is formally defined as follows.

Definition 3: ECA rule. A ECA rule $R_i$ is defined as the implication of the form $Trig \rightarrow A$ where:

- $Trig$ is referred to as triggers which is a set of conditions and/or events connected by the logic operators AND.
- $A$ is referred to as actions which is a set of IoT service functionalities to be executed. Multiple actions can be connected by the logic operators AND.

It is possible that multiple trigger conditions (resp. actions) are connected by the AND operator to create a more complex rule. However, we find that the majority of ECA rules in real-world have only one trigger condition and one action [13]. To align with the real-world ECA rules, we assume that each rule has one trigger condition and one action.

Context is any information that can be used to characterize the situation of a person [14]. In our work, we identify two types of contexts that are relevant to the task of conflict detection.

Definition 4: Current context. The current context, denoted as $CC$, is defined as $CC = <EnvPre, OutEnv>$ where:

- $EnvPre$ refers to the resident's preferred environment state. It is represented as $EnvPre = <env_i, [T_s, T_e], loc>$ meaning that the resident wants the environment variable $env_i$ to keep a state from the start time $T_s$ to the end time $T_e$ at the location $loc$. For example, "*(temperature = 25C, [2pm, 4pm], bedroom)*" describes the resident wants the bedroom temperature to be 25C from 2pm to 4pm.
- $OutEnv = <env_i, [T_s, T_e]>$ refers to the average outdoor environment state from the start time $T_s$ to the end time $T_e$. For example, the average outdoor temperature is 27C.

## III. GENERIC KNOWLEDGE GRAPH CONSTRUCTION AND PROFILING

We propose a new approach to construct the generic knowledge graph $KG$ for IoT services based on the information source ConceptNet. A knowledge graph $KG$ is defined as follows.

Definition 5: Knowledge graph. A knowledge graph $KG$ is a set of triplets $KG = (h, r, t)$. Each triple is composed of a head entity $h \in E$, a tail entity $t \in E$, and a relation $r \in R$ between them (e.g., AC, isUsedFor, cool a room). Here, $E$ denotes the set of entities, and $R$ denotes the set of relations.

Given a set of head entities $H$ (i.e., IoT services) and tail entities $Envir$ (i.e., eight environment entities), our task is to find the relation $r \in \{increase, decrease\}$ between a head entity $h \in H$ and a tail entity $envir \in Envir$. In our work, we employ the knowledge graph ConceptNet for extracting the *increase/decrease* relation. The process of extracting the *increase/decrease* relation is shown in Fig. 4.

### A. Acquiring knowledge

ConceptNet is the knowledge graph for common sense knowledge based on the most basic things a person knows. It connects words and phases of natural language with labeled, weighted edges [15]. Its knowledge is collected from different sources such as Wordnet and DBPedia. So far, it contains over 21 million edges and over 8 million nodes. There are 36 types of relations for connecting nodes. In this paper, we use the *isUsedFor* and *isCapableOf* relations to extract knowledge because they contain the majority of information for describing the IoT services' functionalities and IoT services' impacts on the environment entities. ConceptNet provides a REST API [4] to access data.

Next, we preprocess the extracted raw texts regarding IoT services from ConceptNet. We perform three steps to preprocess the raw texts, namely, *stop word removal*, *Lemmatization*, and *word embedding*. The first step is to filter out stop words that are commonly used but have little importance in the meaning of a sentence. For example, articles (e.g., "the" and "a"), prepositions (e.g., "of" and "to"), and pronouns (e.g., "that" and "this" ) are common stop words. We use the stop words list provided by the NLP toolkit NLTK to remove stop words from the raw texts . The next step focuses on lemmatization. Texts sometimes use different forms of a word (e.g., organize, organizes, and organizing) with similar meanings. The lemmatization step aims to reduce inflectional forms of a word to a common base form. For example, the words "car, cars, car's, cars' " could be converted to the base word "car" after lemmatization. We use the lemmatizer provided by the NLP toolkit NLTK for the second step. Finally, we employ the pretained model BERT to perform the word embedding task. Word embedding is the process of mapping a text to a vector of real numbers for the purpose of computing semantic similarities between words. The BERT model (Bidirectional Encoder Representations from Transformers) is a language representation model which is designed to pretrain deep bidirectional representations from unlabeled text by jointly conditioning on both left and right context [16]. Pretrained representations can also either be contextfree or contextual, and contextual representations can further be unidirectional or bidirectional. Context-free models such as word2vec or GloVe generate a single "word embedding" representation for each word, so the word "bank" would have the same representation in "bank deposit" and "river bank". Contextual models instead generate a representation of each word that is based on the other words in the sentence. Therefore, we use BERT model to embed words for the purpose of computing cosine similarity in the following task.

---

[4] api.conceptnet.io

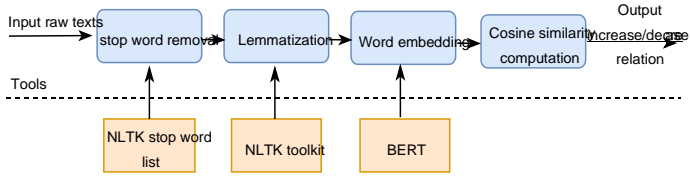

Fig. 4. Workflow of constructing generic knowledge graph

*B. Relation identification*

This section focuses on identifying *increase/decrease* relations interlinking entities. The *increase/decrease* relation (denoted as +/−) indicates an IoT service could change an environment property. We learn the *increase/decrease* relations from the acquired IoT services' knowledge. To learn *increase/decrease* relations between IoT services and environment properties, we first manually build a corpus for each environment property to augment representative information relevant to the conception of the environment property. For each environment variable *envir*, we create two subcorpus, denoted as *Inset* and *Deset*, to incorporate phases that can indicate an *increase* and *decrease* in the value of the environment variable *envir*, respectively. For example, "*hot, increase temperature, hotness, heat, warm, warmth, high temperature*" is likely to indicate an *increase* in the temperature. "*cold, coldness, decrease temperature, low temperature, cool*" is more likely to indicate *decrease* in the temperature.

We identify the *increase/decrease* relations through computing the cosine similarity between the acquired IoT service's knowledge and eight environment properties' corpus by
$$cosine(s_i, envir_j) = \frac{V_i \cdot W_j}{||V_i|| \cdot ||W_j||} \quad (1)$$
Equation (1).

where $V_i \in V$ and $W_j \in Inset \cup Deset$ are word vectors for the extracted knowledge regarding an IoT service $S_i$ and the environment property $envir_j$, respectively. If the similarity value is greater than a threshold $\theta$, the flag between an IoT service and an environment property is signed to be "+" or "-". For example, AC has the functionality of "cool a room". The cosine similarity between "cool a room" and "temperature" is equal to 0.713. Therefore, there is a *decrease* relation between AC and temperature. Algorithm 1 shows the details of finding *increase/decrease* relation between IoT services and environment properties. In this paper, $\theta$ is found to be 0.6.

*C. Context-based Knowledge Graph Profiling*

As per the previous discussion, we construct a seed knowledge graph that is generic coarse-grained and applicable and reusable for all smart homes. For a particular smart home, the seed knowledge graph should be profiled to suit the purpose of conflict detection. Knowledge graph profiling aims to refine the seed knowledge graph to suit the application purpose based on the particular smart home setting and real-time environment property values. There are three subtasks involved in the process of generic knowledge graph profiling as shown in Fig.5. 1) *Context-based knowledge graph completion*

The seed knowledge graph for IoT services contains information including IoT service entities, environment entities, and the *increase/decrease* relations between the IoT services and environment entities. However, the seed knowledge graph lacks the relations between the environment entities and the IoT services. We refer to such type of relations as *trigger* relations. The knowledge graph completion aims to add missing information to the graph [17]. In this section, the task of knowledge graph completion aims to add missing *trigger* relations between the environment entities and the IoT service entities. We extract the *trigger* relations from the ECA rules developed by the resident in a particular smart home. To extract the *trigger* relations, we first extract the environment property $env_k$ and/or the IoT service $S_i$ involved in the trigger condition from the ECA rule $R_k$. We extract the IoT service $S_j$ involved in the action. For example, the *trigger* relation extracted from the rule "temperature>25C → close window" is represented as "(temperature, T, window)" where the flag "T" denotes the *trigger* relation. The ultimate knowledge graph after adding *trigger* relations is denoted as $KG_1$.

Algorithm 1 Find increase/decrease relations
---
Input: *Envir* =< *envir,Inset,Deset* >(a tuple of corpus); D = {$D_1,D_2,...,D_n$} (a set of word vectors)
Output: R (a set of relations)
1: R = Null
2: for $D_i$ in D do
3:     for $w_i$ in *Inset* do
4:        if $consine(D_i,w_i) > \theta$ then
5:           $flag = +$
6:           $R \leftarrow (s_1, envir, flag)$
7:        end if
8:     end for
9:     for $w_j$ in *Deset* do
10:        if $consine(D_i,w_j) > \theta$ then
11:           $flag = -$
12:           $R \leftarrow (s_1,envir,flag)$
13:        end if
14:     end for
15: end for
16: return R
---

2) *Context-based knowledge graph tailoring*

In a real-world smart home context, whether two IoT services impact the shared environment highly depends on their locations. For example, opening the window in the living room has little effect on the AC in the bedroom. Since the seed knowledge graph is learned from commonsense knowledge from different people, the location of the same IoT service might be different in different smart homes. For example, the TV may be placed in the living room in one smart home and is in the bedroom in another smart home. In this regard, we tailor the knowledge graph $KG_1$ based on IoT services' locations. The IoT services in the same location are grouped together and constitute a subknowledge graph. The tailored knowledge graph is denoted as $KG_2$.

3) *Context-based knowledge graph refinement*

The seed knowledge graph contains ambiguous IoT service *impact* relations between some IoT service entities and environment entities. For example, as shown in Fig.2.(a), opening the window may increase or decrease the room temperature. The cause for the ambiguity is that the seed knowledge graph is constructed based on commonsense knowledge provided by different people. The IoT service impact relation between a pair of an IoT service and an environment property might be different based on different people's perceptions. In this regard, the task of knowledge graph refinement aims to disambiguate the IoT service impact relation given a specific context. Here the context includes the resident's indoor environment preferences *EnvPre* and outdoor environment state *OutEnv*. Specifically, the resident's preference towards an environment property can be inferred from ECA rules that have been executed. For example, if the rule R6 ("*temperature > 25C → turn on AC AND temperature = 24C*") is executed and the AC is in use from 2pm to 4pm, we can infer that the resident wants the temperature to keep in a 24C state during this time period. If the average outdoor temperature is more than 24C from 2pm to 4pm, we can infer that opening the window would let hot air come in and *increase* the indoor temperature. Therefore, we can determine the relationship between the window and the temperature is *increase* (denoted as <window, +, temperature>). The refined knowledge graph is denoted as $KG_3$.

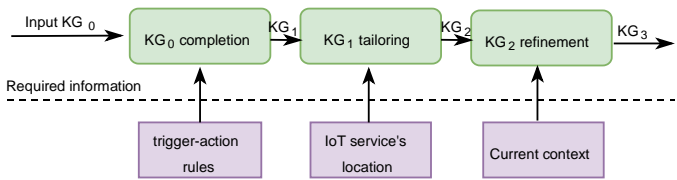

Fig. 5. Workflow of profiling the generic knowledge graph

## IV. A TAXONOMY FOR CONFLICTS

In this section, we formally define four types of basic conflicts. A conflict is manifest when multiple IoT services are invoked at the same time period and in the same location. We first introduce the notion of the time interval-based event for IoT services. We formally define four types of basic conflicts based on the IoT service interval-based event.

Definition 6: Interval-based event. An interval-based event for an IoT service $S_i$ is defined as $e_i =< S_i, [st,et], loc >$ meaning that the IoT service $S_i$ is in use from the start time $st$ to the end time $et$ in the location $loc$. For example, an interval-based event *(AC, 2pm, 4pm, bedroom)* describes that the AC is in use from 2pm to 4pm in the bedroom.

Definition 7: Function-function conflict. A function-function conflict, denoted as *FFConf*, may arise when two ECA rules (i.e., $R_i$ and $R_j$) invoke a shared IoT service functionality. The following condition is satisfied.

$$(Trig_i = true) \wedge (Trig_j = true) \wedge (A_i \neq A_j) \quad (2)$$
$$T_i - T_j < \zeta \quad (3)$$

where $A_i$ and $A_j$ are the IoT service functionality corresponding to the rule $R_i$ and $R_j$, respectively. $Trig_i$ and $Trig_j$ are the trigger condition corresponding to the rule $R_i$ and $R_j$, respectively.

Definition 8: Opposite-environment-impact-conflict. An opposite-environment-impact-conflict (*OppConf*) happens if two IoT services impact the shared environment variable $env_i$ in an opposite way during a time period and in the same location. The following conditions are satisfied.

- $e_i.st \leq e_j.st \leq e_i.et$, meaning that $S_i$ and $S_j$ are invoked simultaneously and there is a temporal overlap between them.
- $e_i.loc = e_j.loc$, denoting that the two IoT services are executed in the same location.

$$env_i = env_j \quad (4)$$

  which means that the two IoT services influence a shared environment variable.

- The two IoT services change the shared environment variable in an opposite way when the following condition is satisfied.
$$eff_i \neq eff_j \quad (5)$$

Definition 9: Cumulative-environment-impact-conflict. A cumulative-environment-impact-conflict(*CumConf*) happens if two IoT services impact the shared environment variable $env_i$ in the same way during a time period and in the same location. The following conditions are satisfied.

- $e_i.st \leq e_j.st \leq e_i.et$, meaning that $S_i$ and $S_j$ are invoked simultaneously and there is a temporal overlap between them.
- $e_i.loc = e_j.loc$, denoting that the two IoT services are executed in the same location.

$$env_i = env_j \quad (6)$$

  which means that the two IoT services influence a shared environment variable.

- The two IoT services change the shared environment variable in the same way when the following condition is satisfied. $eff_i = eff_j$ (7)

Definition 10: Transitive-environment-impact-conflict. A transitive-environment-impact-conflict (*TraConf*) happens if the impact of an IoT service on the environment variable $env_i$ triggers the execution of another IoT service accidentally. The following conditions are satisfied.

- $e_i.st \leq e_j.st$, meaning that $S_i$ is invoked before $S_j$.
- $e_i.loc = e_j.loc$, denoting that the two IoT services are executed in the same location.
- The impact of the IoT service $S_i$ on the environment variable $env_i$ makes the IoT service $S_j$'s trigger condition become true, that is,

$$Trig_j = true \qquad (8)$$

## V. CONFLICT DETECTION ALGORITHM

We develop the algorithm Conflict-detector to efficiently detect conflicts based on our taxonomy. The Conflict-detector consists of the *finding overlapping event pairs* module and the *conflict detection* module. Each of the modules is explained as follows.

Algorithm 2 focuses on finding time interval-based event pairs that have temporal overlap. We sort all elements in $E$ based on their start time to reduce the searching scope. If two IoT services have time interval overlap, we collect them and insert them to the set $O$ (Algorithm 2 line 3-7).

Next, we aim to detect conflicts using Algorithm 3. Given a set of executed ECA rules and time constraint $\zeta$, we detect whether there are function-function conflicts based on Definition 7 (Algorithm 3 line 2-3). For any interval-based event pairs $<e_i,e_j>$, the corresponding IoT services for $e_i$ and $e_j$ are $s_i$ and $s_j$, respectively. We find the set of triplets from $KG_3$ whose tail or head is $s_i$ (denoted as $Set_i$). Similarly, we can find $Set_j$ for $s_j$ (Algorithm 3 line 4-8). Then, we detect the conflicts between two triplets $\alpha$ and $\beta$ based on Definition 8-10 (Algorithm 3 line 9-15).

---

**Algorithm 2** Find overlapping event pairs algorithm

Input: $E = \{e_1,e_2,...,e_m\}$(a set of time interval-based events)
Output: $O$ (a set of overlapping interval-based events)
1: $O$ = Null
2: $E'$= sorting($E$)// sort elements in $E$ based on event's start time.
3: for $e_i$ in $E'$ do
4:    for $e_j$ in $E'$ do
5:      if $e_i.st < e_j.st < e_i.et$ and $e_i.loc = e_i.loc$ then 6:   $O \leftarrow <e_i,e_j>$//find a pair of overlapping events and insert them to $O$.
7:      end if
8:    end for
9: end for
10: return $O$

---

**Algorithm 3** Conflict detection algorithm

Input: $O$ (a set of overlapping interval-based events); $KG_3$ (a knowledge graph profile); $\zeta$(time constraint)
Output: $ConfSet$(a set of conflicts)
1: $ConfSet$ = Null
2: if $(Trig_i = true) \wedge (Trig_j = true) \wedge (A_i = A_j) \wedge (T_i - T_j < \zeta)$ then
3:   $ConfSet \leftarrow (R_i,R_j,FFConf)$
4: for $<e_i,e_j>$ in $O$ do
5:   for item in $KG_3$ do
6:     if item.h == $s_i \vee$ item.t ==$s_i$then$Set_i \leftarrow$ item
7:     if item.h == $s_j \vee$ item.t ==$s_j$then$Set_j \leftarrow$ item
8:   end for
9:   for $\alpha$ in $Set_i$ do
10:     for $\beta$ in $Set_j$ do
11:       if $\alpha.t == \beta.t$ and $\alpha.r == \beta.r$ then $ConfSet \leftarrow (\alpha,\beta,$ Cumconf$)$
12:       if $\alpha.t == \beta.t$ and $\alpha.r \, != !\beta.r$ then $ConfSet \leftarrow (\alpha,\beta,$ Oppconf$)$
13:       if $\alpha.t == \beta.h$ and $\beta.r ==$T then $ConfSet \leftarrow (\alpha,\beta,$ Tranconf$)$
14:     end for
15:   end for
16: end for
17: return $ConfSet$

---

## VI. EVALUATION

We conduct a set of experiments to evaluate our proposed approach. Experiments are performed on a 3.4 GHZ Intel processor and 8 GB RAM under Windows 10. We evaluate our proposed approach using three metrics including *efficiency*, *scalability*, and *performance* in terms of precision, recall and F1 score which are defined as follows.

$$precision = \frac{TP}{TP+FP} \qquad (9)$$

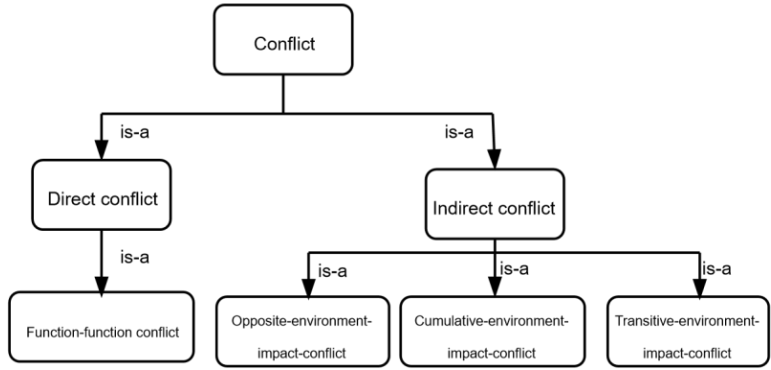
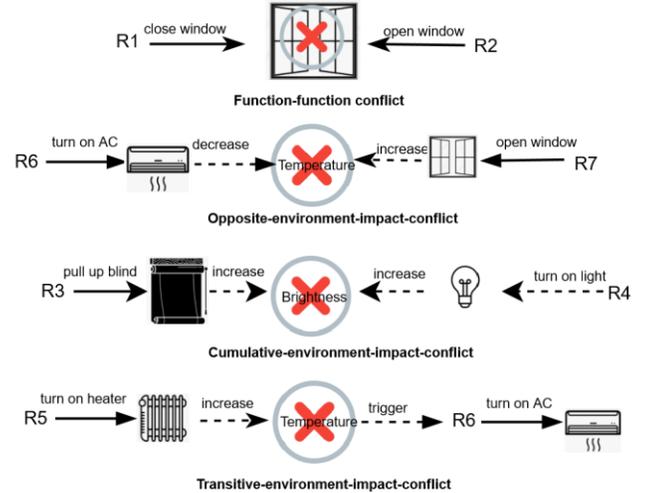

(a)          (b)

Fig. 6. A taxonomy for conflicts

$$recall = \frac{TP}{TP + FN} \quad (10)$$

$$F1 = 2 * \frac{precision * recall}{precision + recall} \quad (11)$$

where *TP* is the number of true positives, *FP* is the number of false positives, and *FN* is the number of false negatives.

*A. Experiment set up and dataset description*

We test the efficiency, scalability, and performance on both synthesized datasets and a real-world dataset. The real-world dataset and the process of synthesizing datasets are described as follows.

*Real-world datasets*: The real-world dataset was collected from an old person who lived in a single apartment [18]. There were 84 sensors attached on daily life things such as doors, windows, cabinets, drawers, microwave ovens, refrigerators, stoves, sinks, toilets, showers, light switches, lamps, some containers (e.g water, sugar, and cereal), and electric/electronic appliances (e.g DVDs, stereos, washing machines, dishwashers, coffee machines). The locations of these sensors are also provided. The start time and end time of related sensors are recorded when the resident performs activities. Fig.7 shows some samples of the dataset. For example, the "taking medication" is performed from time 04:23:06am to 04:41:32am in the kitchen on the date of 5/3/2003. There are 6 things are involved in this activity including Light switch with sensor ID 75, Door with sensor ID 51, Sink faucet with sensor ID 91, etc. Each thing has a start time and an end time. The Light switch identified by sensor 75 starts at 04:30:23am and ends at 04:33:39am. A total of 21 activity types are annotated and recorded. All these activities are performed in a natural setting without any interruption in the daily life of the resident. The data was collected for two weeks.

*Synthesized datasets*: The setting of generating datasets is shown in TABLE II. We create a range of 10 to 50 numbers of smart homes. For each smart home, there are 130 numbers of things involved which are collected from the real smart home [18] and the website IFTTT. A number of 300 interval-based events are generated for each smart home during a day. For each interval-based event, its start time is between 00:00-23:59. The duration of the event is between 60 minutes and 120 minutes. Each smart home has five rooms including a kitchen, a living room, a bathroom, a bedroom, a study room. We consider 7 numbers of common indoor environment entities. The value of each environment entity is a random integer generated from a range. For example, the indoor temperature in a particular smart home is randomly generated ranging from 22 to 28. For simplicity, we also generate a resident's preferred environment value using the same setting. There are 277 numbers of conflicts which are used as the ground truth.

TABLE II
SETTING OF SYNTHESIZING DATASETS

| Variable | Value or range |
| --- | --- |
| No. of smart homes | 10,20,30,40,50 |
| No. of things | 130 |
| No. interval-based events | 300 |
| Start time of interval-based events (st) | 00:00-23:59 |
| End time of interval-based events(et) | [st + 60, st+120] |
| Location | Kitchen, living room, bathroom, bedroom, study room |
| Indoor temperature | 22-28 |
| Indoor humidity | 45-55 |
| Indoor $CO_2$ | 350-450 |
| Indoor brightness | 60-70 |
| Indoor sound | 10-70 |
| Indoor smoke | 0-100 |
| Indoor ventilation | (0-3) |
| No. of conflicts | 277 |

*B. Experiment results and analysis*

*1) Experiments on real-world datasets:* We conduct the first experiment on the real-world datasets to evaluate the efficiency of our conflict detection algorithm. Fig.9 shows the efficiency of the conflict detection algorithm in terms of execution time. For example, the execution time of the conflict-detection algorithm on the first day is 272ms. The average execution time for the 14 days is 255ms. On the 10th and 13th day, the execution time is marginally larger than the average execution time because the resident performs more activities on the two days than the rest of the days. The efficiency indicates that our algorithm is suitable for real-time conflict detection in the smart home environment.

We conduct the second experiment on the real datasets to evaluate the scalability of our conflict detection algorithms. Since there are no conflicts recorded in this real data, we use execution time with respect to the numbers of interval-based events to evaluate the scalability. The experiment result is shown in Fig.8. We vary the numbers of time interval-based events from 100 to 300 to test the execution time of the conflict detection algorithm. All experiments are conducted 10 times and the average execution time is computed. For example, the average execution is 258ms when there are 100 interval-based events. The experiment result shows that the execution time of the conflict detection algorithm linearly increases as the size of interval-based events grows.

| Taking medication | 5/3/2003 | 4:23:06 | 4:41:32 | Kitchen | | |
| --- | --- | --- | --- | --- | --- | --- |
| 75 | 51 | 91 | 96 | 119 | 74 | |
| Light switch | Door | Sink faucet | Sink faucet | Light switch | Refrigerator | |
| 4:30:23 | 4:30:34 | 4:31:29 | 4:31:33 | 4:32:10 | 4:32:40 | |
| 4:33:39 | 19:41:00 | 4:55:53 | 6:11:38 | 20:30:34 | 4:32:47 | |
| Preparing breakfast | 5/3/2003 | 5:29:51 | 6:24:47 | Kitchen | | |
| 54 | 74 | 75 | 74 | 108 | 84 | 108 |
| Cabinet | Refrigerator | Light switch | Refrigerator | Toaster | Garbage disposal | Toaster |
| 5:36:36 | 5:37:23 | 5:37:55 | 5:38:08 | 5:38:27 | 5:39:54 | 5:40:20 |
| 5:36:38 | 5:37:50 | 5:40:31 | 5:38:24 | 5:38:33 | 6:11:57 | 6:15:37 |

Fig. 7. Samples of the real-world datasets

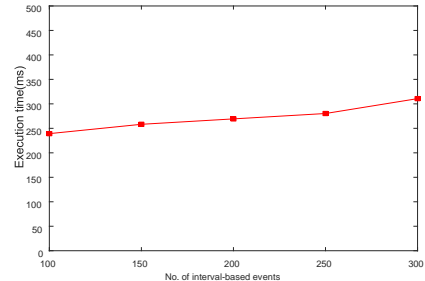

Fig. 8. Scalability of conflict detection on real datasets

We conduct the third experiment on the real dataset to evaluate the performance of our conflict detection approach. Since there are no conflicts recorded in the real dataset. In this regard, we augmented the real dataset by introducing more things that have impacts on environment. We set the time constraint $\zeta$ to be 5 minutes meaning that two ECA rules invoke the shared IoT service within 5 minutes. For example, one rule requests to turn on the light at 12:00pm. Another rule requests to turn off the light between 12:00pm to 12:05pm. Since there is no ground truth for conflicts, the results are analyzed by two experts. We trained the experts by demonstrating the conflicts in the motivating scenario to enable them have basic knowledge about conflicts. In this regard, the experts can provide external labels for each result based on our conflict taxonomy and their perceptions about conflicts. The performance in terms of precision, recall, and F1 score of conflict detection is illustrated in Fig.10. We can see that our approach can find all conflicts when the time constraint threshold $\zeta$ is set to be 5 minutes. For example, the precision, recall, and F1 score are 0.81, 1, and 0.89, respectively. We analyze the false positive scenarios and provide two filtering strategies to improve our approach.

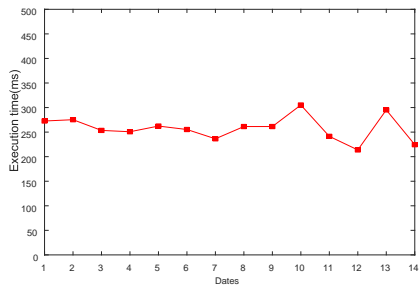

Fig. 9. Efficiency of conflict detection on real datasets

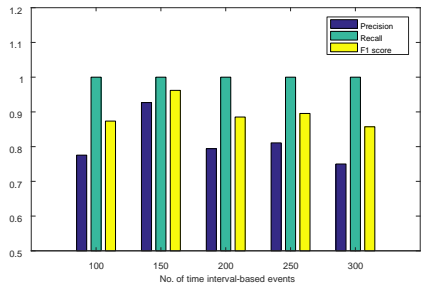

Fig. 10. Performance of conflict detection approach on real datasets

*2) Filtering strategies:* Since most things such as "cabinet" and "drawer" have limited impact on the environment, many interval-based events play a little role in conflict detection. For example, using table and chair has little impact on any environment entities. In this regard, it is feasible to filter out these events before detecting conflicts.

Additionally, some non-conflicting event pairs are considered as conflicts because the closed scope of effect of an IoT service is not considered. For example, a false positive scenario is the event pair "(AC, kettle)" which is detected as an oppositeenvironment-conflict. The possible reason is that the property of the kettle (e.g.,"heating water" ) is considered to be most relevant to increase the temperature by our method. As a result, there is a relation between the kettle and the temperature ("kettle, +, temperature"). Once the AC and the kettle are used simultaneously, they will be considered to have a conflict over the temperature. Other false positive scenarios such as "(AC, iron)" and "(fridge, microwave)" are contributed to the same reason that the impact scope of the IoT services is ignored.

Based on the above discussion, we propose two filtering strategies to further improve our approach. The first strategy is to filter out the interval-based events whose involved things have little impact on the environment. We use our knowledge graph as a reference to categories things into two groups. Things that occur in both the knowledge graph and a smart home are considered for conflict detection. The rest of things are filtered out and are not considered. The second strategy is to further filter out non-conflicting event pairs using contexts. We consider two type of contexts including the current indoor environment value and the resident's own preferred environment value. For example, whether opening the window and turning on the AC is a conflict highly depends on the current indoor temperature and the resident's preferred temperature. If the indoor temperature is 22C and the resident's preferred temperature is also 22C, opening the window and turning on the AC simultaneously should not be considered as a conflict. If the indoor temperature is 25C while the resident's preferred temperature 22C, a conflict may occur because opening the window may let hot air in when the AC is cooling.

*3) Experiments on synthesized datasets:* : We test the effectiveness of the two filtering strategies in conflict detection. In the following, we refer our proposed conflict detection approach as M1. Our conflict detection approach combined with strategy one is referred to as M2. Our conflict detection approach combined with strategy two is referred to as M3. Our conflict detection approach combined both strategies is referred to as M4. The experiment setting is shown in TABLE II. We conduct the experiment on different numbers of smart homes.

We conduct the fourth experiment on the synthesized datasets to test the performance of the four methods. We evaluate the performance in terms of precision, recall, and F1 score based on different numbers of smart homes ranging from 10 to 40. The experiment result is shown in Fig.11. For example, the precision, recall, F1 score of the method M4 which is running on 10 numbers of smart homes are 0.72, 1, and 0.84, respectively. The experiment result in Fig.11 shows that the overall performance of our conflict detection approach combined with two strategies (M4) outperforms M1, M2, and M3 methods.

We conduct the fifth experiment on the synthesized datasets to evaluate the efficiency of the four methods. The efficiency in terms of average execution time is shown on the left side of Fig.12. It is clear that the average execution time linearly increases as the numbers of smart homes increase. It is an expected result because the larger numbers of smart homes, the larger size of the input datasets. We also see that the two filtering strategies can slightly improve the algorithm's efficiency. When our conflict detection algorithm combined with the two strategies (M4), the efficiency is greatly improved because the execution time is smaller than M1, M2, and M3. We conduct the sixth experiment to test the scalability of the four methods under different numbers of smart homes. As shown in Fig.12, the numbers of detected conflicts increases as an increase in the numbers of smart homes. We can also see that the method M4 discover the least numbers of conflicts because applying the two strategies can greatly reduce the search space and remove non-conflicting event pairs.

## VII. RELATED WORK

In the smart home setting, detecting conflicts for appliance interactions is a hot problem. In [19], a taxonomy for the interactions between networked appliances is presented for better understanding the problem. Four types of interactions are provided including multiple action interactions, shared trigger interactions, sequential action interactions, and missed trigger interactions. These interactions are categorized based on the situation in which the interaction occurs within a single service or among multiple services. In [20], an object-oriented approach for modeling appliances and the environment is proposed. The appliances are modeled as a set of methods and attributes. The environment such as the temperature is modeled as a property. Two types of interactions are modeled including appliance interactions and environment interactions. In [9], the side effects of appliances are considered in modeling and detecting appliance interactions. For example, an AC cools the air as its main function but also decreases the humidity as a side-effect. In [8], an object-oriented approach is presented to model appliances as a set of methods and properties. Two types of interactions are modelled including appliance interactions and environment interactions. In [10], a rule-based approach is proposed to detect the conflicts among services. Users' preferences are specified using rules which consist of user, triggers, environment, and actuators attributes. Conflicts among appliances can be identified through analyzing the rule relations. A framework RemedIoT is proposed to detect the direct conflicts among IoT devices. A a set of policies, are devised to define the allowable and restricted state-space transitions of IoT devices [7]. A conflict detection approach at design time is proposed in [9]. A complete flow technique is devised to identify and resolve direct actuation conflicts for IoT-based applications

## VIII. CONCLUSION AND FUTURE WORK

We propose a conflict detection framework in IoT-based smart homes. We employ knowledge graph to capture the IoT services' impact on environment entities. We also profile the generic knowledge graph to the context-based knowledge graph based on the current indoor environment and the resident's preferences. A taxonomy for conflicts is proposed and four types of conflicts are modelled. We propose a novel algorithm to detect conflicts efficiently and two filtering strategies are proposed. The experimental results demonstrate the performance and efficiency and scalability of our proposed framework. The novelty of our work is that we can detect conflicts automatically without human intervention. Future work intends to focus on resolving conflicts.

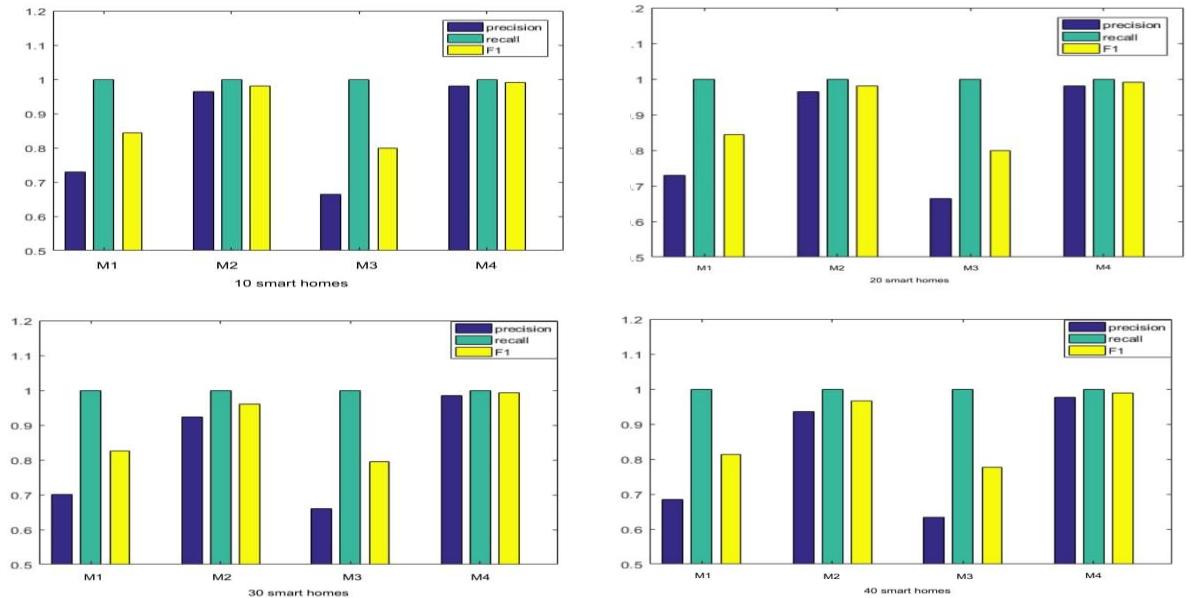

Fig. 11. Performance of filtering strategies on synthesized datasets

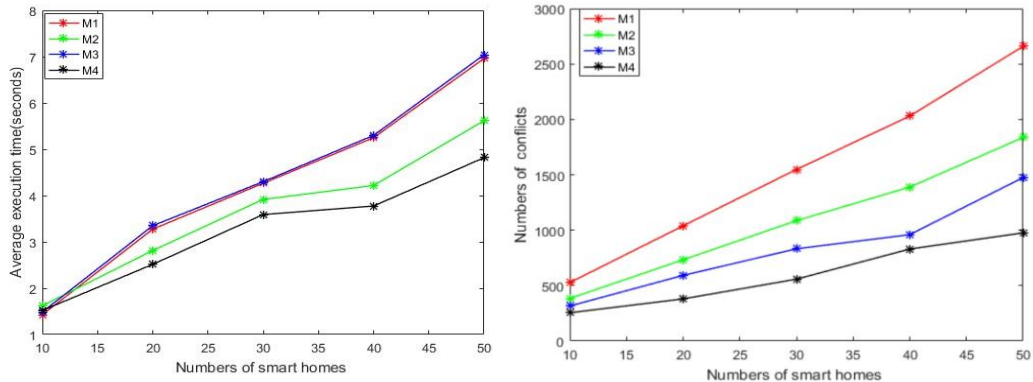

Fig. 12. Efficiency and Scalability filtering strategies on synthesized datasets

ACKNOWLEDGEMENT

This research was partly made possible by DP160103595 and LE180100158 grants from the Australian Research Council. The statements made herein are solely the responsibility of the authors.